\documentclass[letterpaper, 10 pt, conference]{ieeeconf}  % Comment this line out if you need a4paper
\usepackage{cite}
\usepackage{amsmath,amssymb,amsfonts}
\usepackage{siunitx}
\usepackage{algorithmic}
\usepackage{graphicx}
\usepackage{textcomp}
\usepackage{graphicx}
\usepackage{multicol}        % used for the two-column index
\usepackage{multirow}

\PassOptionsToPackage{hyphens}{url}\usepackage{hyperref}
\usepackage[table]{xcolor}
\def\BibTeX{{\rm B\kern-.05em{\sc i\kern-.025em b}\kern-.08em
    T\kern-.1667em\lower.7ex\hbox{E}\kern-.125em}}
\begin{document}

\title{
%Robotic Vision to Ensure Safety \\ in %Human-Robot Collaboration Scenarios\\ Barrierless Safety System for Robotic Manipulators Based on Computer Vision\\ Safety System for Barrierless  Human-Robot Collaboration
%Safety System with Robotic Vision for Barrierless  Industrial Manipulator\\
%Safety System for Barrierless \\Industrial Manipulators\\
Vision-Based Safety System for\\
Barrierless Human-Robot Collaboration
%\thanks{Identify applicable funding agency here. If none, delete this.}
}

% \author{\IEEEauthorblockN{Lina María Amaya Mejía}
% \IEEEauthorblockA{\textit{dept. Electronic Engineering} \\
% \textit{Faculty of Engineering} \\
% \textit{Pontificia Universidad Javeriana}\\
% Bogotá, Colombia \\
% linaamaya@javeriana.edu.co}
% \and
% \IEEEauthorblockN{Nicolás Duque Suárez}
% \IEEEauthorblockA{\textit{dept. Electronic Engineering} \\
% \textit{Faculty of Engineering} \\
% \textit{Pontificia Universidad Javeriana}\\
% Bogotá, Colombia \\
% n-duque@javeriana.edu.co}

% \and
% \IEEEauthorblockN{Daniel Jaramillo Ramírez}
% \IEEEauthorblockA{\textit{dept. Electronic Engineering} \\
% \textit{Faculty of Engineering}\\
% \textit{Pontificia Universidad Javeriana}\\
% Bogotá, Colombia \\
% d-jaramillo@javeriana.edu.co}

% \and
% \IEEEauthorblockN{Carol Viviana Martínez Luna}
% \IEEEauthorblockA{\textit{SnT -  Interdisciplinary Centre} \\ 
% \textit{for Security, Reliability and Trust} \\
% \textit{University of Luxembourg}\\
% Luxembourg, Luxembourg \\
% carol.martinezluna@uni.lu}
% }

\author{Lina María Amaya-Mejía\textsuperscript{1} \textsuperscript{2}, Nicolás Duque-Suárez\textsuperscript{1}, Daniel Jaramillo-Ramírez\textsuperscript{1}, and Carol Martinez\textsuperscript{2}}% <-this % stops a space
\maketitle

\footnotetext[1]{Dept. Electronic Engineering, Faculty of Engineering, Pontificia Universidad Javeriana, Bogotá, Colombia
        {\tt \{linaamaya},
        {\tt n-duque},
        {\tt d-jaramillo\}@javeriana.edu.co}%
}
\footnotetext[2]{Space Robotics (SpaceR) Research Group, Interdisciplinary Centre for Security, Reliability and Trust (SnT), University of Luxembourg, Luxembourg. 
        {\tt lina.amaya.001@student.uni.lu},
        {\tt carol.martinezluna@uni.lu}%
}

\maketitle

\begin{abstract}
Human safety has always been the main priority when working near an industrial robot. With the rise of Human-Robot Collaborative environments, physical barriers to avoiding collisions have been disappearing, increasing the risk of accidents and the need for solutions that ensure a safe Human-Robot Collaboration. This paper proposes a safety system that implements Speed and Separation Monitoring (SSM) type of operation. For this, safety zones are defined in the robot's workspace following current standards for industrial collaborative robots. A deep learning-based computer vision system detects, tracks, and estimates the 3D position of operators close to the robot. The robot control system receives the operator's 3D position and generates 3D representations of them in a simulation environment. Depending on the zone where the closest operator was detected, the robot stops or changes its operating speed. Three different operation modes in which the human and robot interact are presented. Results show that the vision-based system can correctly detect and classify in which safety zone an operator is located and that the different proposed operation modes ensure that the robot's reaction and stop time are within the required time limits to guarantee safety.
\end{abstract}

%\begin{IEEEkeywords}
%\end{IEEEkeywords}

\section{Introduction}
Since the first industrial robots appeared, safety for operators working near them has been one of the most critical priorities. Before the 21st century, it was common to see industrial robots separated from people by physical barriers \cite{I1}. However, technological advances in robotics have allowed people to create human-robot collaborative work environments. This concept of collaborative robotics has been regulated by different standards such as the ISO 10218-1/-2, and the ISO/TS 15066 \cite{I2}. Instead of using physical barriers, collaborative robots are intended for direct interaction within a shared workspace. Therefore, they need tighter safety-related specifications that diminish the possibility of an accident. 

ISO/TS 15066 specifies a type of collaborative operation called Speed and Separation Monitoring (SSM), which reduces the collision risk by keeping a minimum distance between the operator and the robot. The collision risk reduction is achieved by supervising the operator's distance from the robot and the robot's speed. If the minimum separation distance is violated, the robot will go into a protective stop\cite{I3}. 

%It has been established the rise in collaborative robotics and the need of these robots to have certain safety measures, that allows them work in the same space with a human despite the lack of a physical barrier. This brings into consideration the need of a safety system capable of detecting when a subject is close to an operating robot, and acting according to the distance between each other, with the main goal of avoiding a collision.

In the literature, different approaches have been proposed to allow a closer and barrierless collaboration between humans and industrial robots. For instance, in \cite{1}, a safety system was developed that used two RGB-D cameras and a laser system to track the operator and decrease the robot's speed or stop it depending on its distance to the operator. \cite{2} mentions a 3D simulated workspace, where the operator was inserted as a point cloud from an RGB-D camera. Different safety modes were developed, such as alarming the operator, stopping the robot, and modifying its trajectory. The authors in \cite{3} proposed a "potential field" around the operator, which was located by using RGB-D cameras.

Continuing the research started by \cite{7}, this paper proposes a safety system that implements the SSM-type of operation. Safety zones are defined in the robot's workspace according to a risk assessment and calculations following current standards. A deep-learning-based computer vision system monitors the zones by detecting and tracking operators that enter the scene and estimating their 3D position. The robot's control system uses this information to lower the robot's speed or to stop the robot, depending on the operator's location. Other operation modes are also implemented, which vary how the human and the robot interact.

%The integration of these systems was using ROS (Robot Operating System) creating a virtual environment, where a simulated robot was found along a representation in the simulated space of real subjects moving around the supposed location of the robot in real-time. A trained Convolutional Neural Network was used to detect the subjects, in a scene that was captured by an RGB-D camera \cite{14}.

%, which are based on the previously defined zones.

%The calculation of the boundaries of the safety zones are made according to the current standards and technical specifications. The proposed safety system is composed by a computer vision system that is in charge of detecting the subjects in the workspace and estimating their 3D position (and safety zone), and a robot's control system that manages the reactions of the robot according to the position and zone of the closest subject. The integration of these systems was made by creating a virtual environment, where a simulated robot was found along a representation in the simulated space of real subjects moving around the supposed location of the robot in real-time. A trained Convolutional Neural Network was used to detect the subjects, in a scene that was captured by an RGB-D camera \cite{14}.

%\section{Related Work}

The novelties of the safety system presented in this paper are: 
\begin{enumerate}
    \item A deep learning strategy to detect and track humans in industrial human-robot collaboration environments.
    
    \item Three different operation modes: An \textit{SSM-based} operation mode that follows the ISO/TS 15066; a \textit{dynamic safety zones} mode, where the reaction of the robot depends on its euclidean distance to the operator; and an \textit{obstacle evasion} mode, where the robot modifies its trajectory to avoid contact with any operator that is on its way. 
    
    \item Additionally, we include the 3D representation of the workspace using ROS (Robot Operating System) and MoveIt!, which allow a direct interaction/communication between the control and the perception systems and give the control system re-planning and collision avoidance capabilities.
    
\end{enumerate}

%The novelties of the safety system presented in this paper are: 
%{\color{red}(1) A deep learning strategy to detect and track humans in industrial human-robot collaboration environments. (2) Three different operation modes: An \textit{SSM-based} operation mode that follows the ISO/TS 15066; a \textit{dynamic safety zones} mode, where the reaction of the robot depends on its euclidean distance to the operator; and an \textit{obstacle evasion} mode, where the robot modifies its trajectory to avoid contact with any operator that is on its way. Additionally, we include (3) the 3D representation of the workspace using ROS (Robot Operating System) and MoveIt!, which allow a direct interaction/communication between the control and the perception systems and give the control system re-planning and collision avoidance capabilities.}

The paper is structured as follows. Section II introduces the problem statement and the risk assessment. Section III describes the barrierless safety system, which includes the definition of the safety zones, the vision system and the operation modes. Section IV presents the experiments and results that the performance of the safety system. Lastly, Section V presents the conclusions and future work.

\section{Problem Statement}

\begin{figure}[h]
\centerline{\includegraphics[scale=.17]{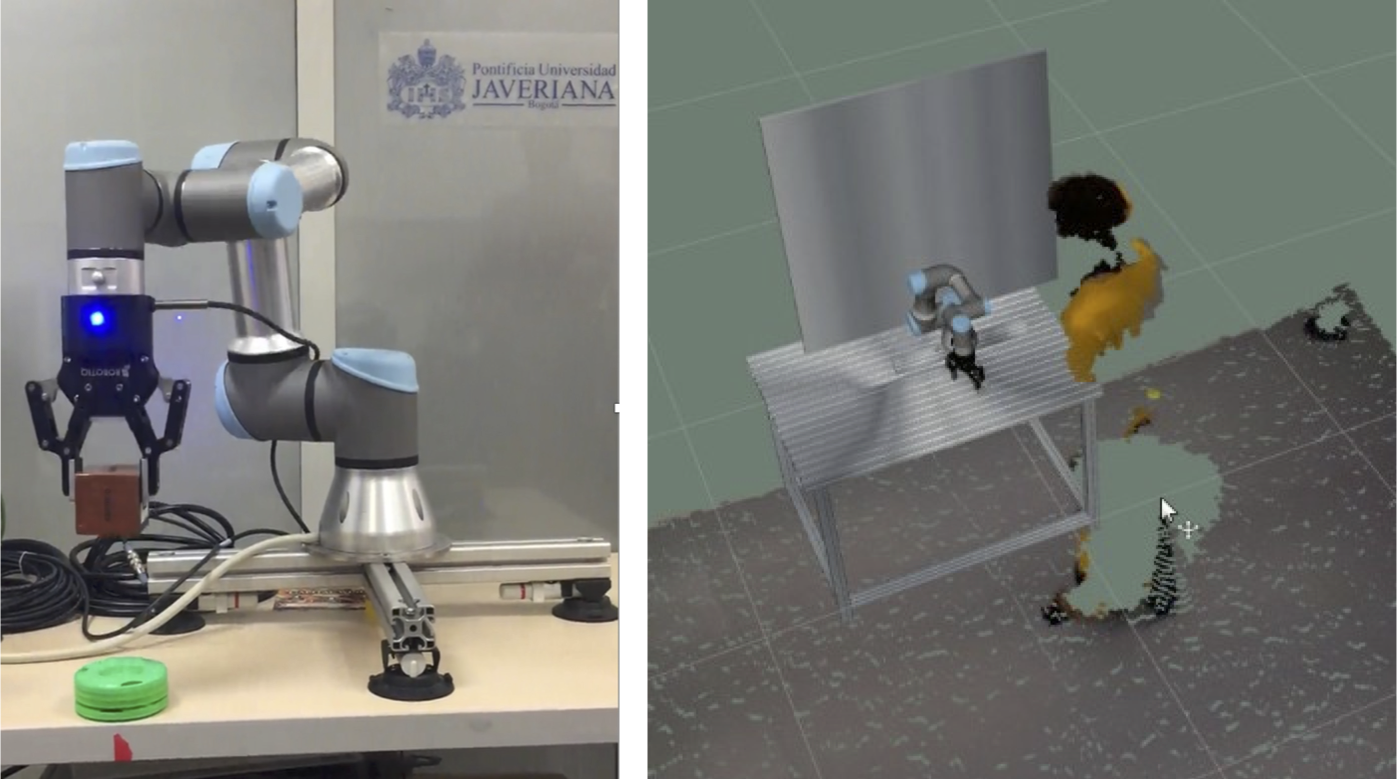}}
\setlength{\abovecaptionskip}{3pt}
\setlength{\belowcaptionskip}{-10pt}
\caption{Barrierless scenario. UR3 performing a pick-and-place routine (left). The robot and operator are sharing the workspace (right). A camera facing downwards monitors the robot-operator interaction.}
\label{intro}
\end{figure}

The system was developed for a scenario where a robotic manipulator performs a pick-and-place routine on a table while operators move around it. A Kinect camera was located at the height of 3.45 m  facing downwards, reaching an area of 4.2~m~x~3.1~m of the scene of interest. Fig. \ref{intro} shows the system's setup. The left side shows the UR3 robotic arm used in the study with a Robotiq 2F-85 gripper. The right side shows an example of what the camera captures when an operator enters the robot's workspace.

%The UR3 It is a 6 DoF industrial robotic arm  
%designed to simulate repetitive manual tasks weighing up 
%, which can weigh up to 3kg within a radius of 500 mm. The end effector is the Robotiq 2F-85 \cite{13}. This is a 2-finger adaptive gripper for collaborative robots, with 85 mm opening. 

%, and present three different operation modes, designed to avoid a collision between a human and a robot, which are based on the previously defined zones.

\subsection{Risk Assessment} \label{risk_as}

%ABB \cite{6} carried out a risk assessment for the design of its collaborative robots,
A risk assessment for the UR3 was carried out based on \cite{6} and following the guidelines of the ISO 10218 and ISO 13849 standards to determine the reliability requirements for risk reduction measures, based on assessing the basic properties of hazard defined in Table \ref{hazard_table}.

\begin{table}[htbp]
\caption{Properties of hazard}
\label{hazard_table}
\begin{tabular}{ll}
\hline
\multicolumn{1}{c}{\textbf{Property of hazard}} & \multicolumn{1}{c}{\textbf{Possible values}} \\ 
\hline
\hline
S - Severity of injury & \begin{tabular}[c]{@{}l@{}}S1 - slight, normally reversible injuries\\ S2 - severe,  normally irreversible injuries\end{tabular} \\ \hline
\begin{tabular}[c]{@{}l@{}}F - Frequency of exposure\\       to hazard\end{tabular} & \begin{tabular}[c]{@{}l@{}}F1 - rare\\ F2 - frequent, constant\end{tabular} \\ \hline
\begin{tabular}[c]{@{}l@{}}P - Possibility of  avoiding \\      hazard or limiting harm\end{tabular} & \begin{tabular}[c]{@{}l@{}}P1 - possible under certain conditions\\ P2 - scarcely possible\end{tabular} \\ \hline
\end{tabular}
\end{table}

From the selection of the properties of hazard, the required safety Performance Level (PL) is obtained and ranked on a scale of increasing degree of risk reduction from PLa to PLe.

\begin{figure}[htbp]
\centerline{\includegraphics[scale=.4]{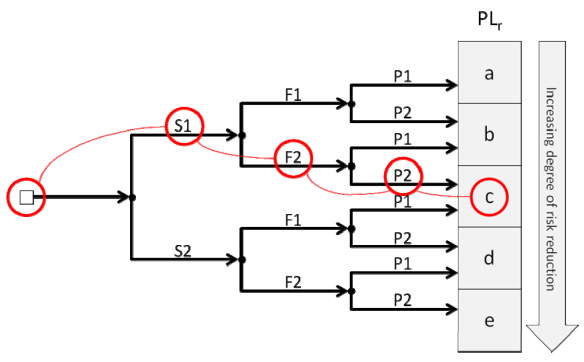}}
\caption{Risk graph for a collaborative robot (ISO 13849) \cite{6.1}}
\label{risk_graph}
\end{figure}

For a collaborative robot like the UR3, the PL is PLc (Fig.~ \ref{risk_graph}). It is a medium level, one level less than the required one for standard industrial robots (PLd). A PLc implies that a medium-level risk reduction measure must be taken using an external safety system. The result is consistent since collaborative robots are designed to interact with humans. They should represent a lower risk than a standard industrial robot. However, as there is a higher frequency of exposure to hazards, other safety measures must be implemented to reduce the risk of injury. Because of this, we propose a vision-based safety system to achieve the required PLc level.

\section{Barrierless Safety System}

%\subsection{General framework}\label{general_fw}
The system was planned under the Speed and Separation Monitoring (SSM) type of operation, in which the robot’s speed depends on the constant evaluation of the horizontal distance of the operator relative to the robot \cite{4}. The distance analysis is performed by a vision system that communicates with the UR3 control system to vary its speed.

\begin{figure*}[htbp]
\centerline{\includegraphics[scale=0.22]{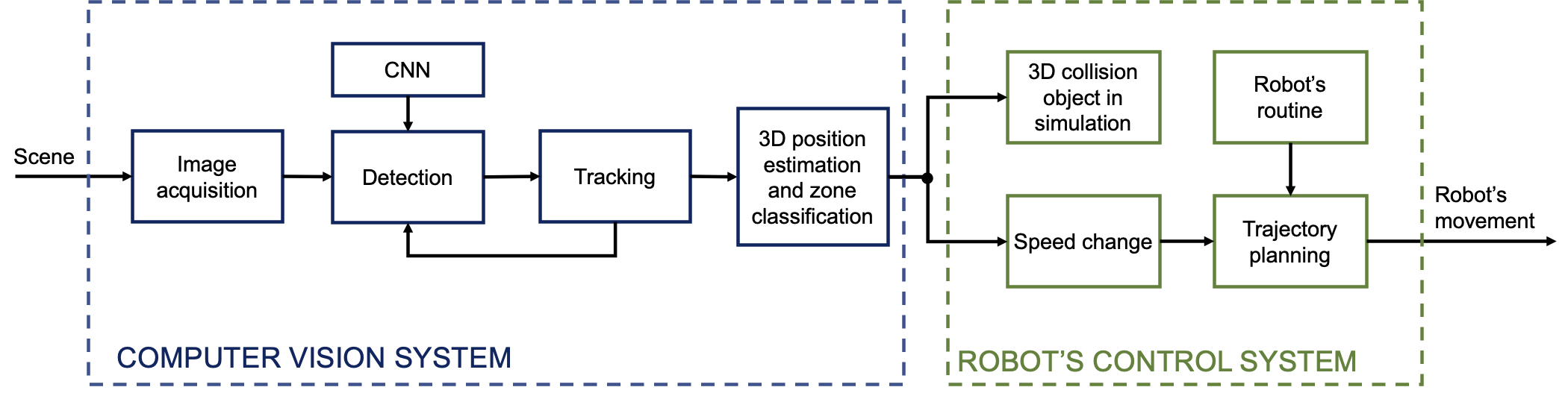}}
\caption{System Architecture. A vision-based system is used to detect, track, and estimate the operator's 3D position and the corresponding safety zone (left side). The robot's control system (right side) represents the operator as a collision object and changes the robot's speed based on the human-robot distance.}
\label{sys_diag}
\end{figure*}

Fig. \ref{sys_diag} presents the architecture of the proposed safety system. The left side describes the vision system where a convolutional neural network (CNN) is used for the detection of the operators nearby the robot, and a tracking algorithm is used to follow them while on scene (see Section~\ref{vision_sys}). The operators' 3D position and height are estimated in real-time and are used to identify the safety zones and the risk of collision (determined by the human-robot distance).

The right side of Fig. \ref{sys_diag} describes the robot's control system. It receives the 3D position and height of the operators and the risk zone where they are detected. This data is then used to generate 3D figures in a simulated space, representing the operators as collision objects to the robot with the color of their respective safety zone. The 3D position is also used to find the human-robot distance and change the robot's speed based on their closeness to avoid a collision. The robot performs a programmed routine that adapts to the speed changes. The robot's routine and other capabilities, like collision avoidance, were possible by taking advantage of the integration of the control system with ROS and MoveIt \cite{19,20}.  
%which provided the necessary tools to make this possible.

% \begin{figure}[htbp]
% \centerline{\includegraphics[scale=.2]{scene.png}}
% \caption{Map of the work scene seen from the camera}
% \label{work_map}
% \end{figure}

\subsection{Definition of Safety Zones}\label{def_zones}
The safety zones around the robot were defined based on the technical specification ISO/TS 15066: 2016 and studies carried out in \cite{8,9}. These studies suggest that it is possible to limit the problem of the minimum separation distance (MSD) between the robot and an operator to a 1D-problem, where the MSD $S_a$ is calculated as follows.

\begin{equation}
S_a= S_h+S_r+S_s+C+Z_d+Z_r
\label{eq1}
\end{equation}

$S_a$ is expressed as the sum of contributions from:
\begin{itemize}
    \item The operator's movement: human distance $S_h$.
    \item The robot's movement: reaction distance $S_r$ and stop distance $S_s$.
    \item The sensors' attributes used for distance measurement: intrusion distance $C$, uncertainties from the robot's position $Z_d$, and the operator's position $Z_r$.  
\end{itemize}

$C$ can be discarded as it is an attribute for devices installed parallel to the robot's z-axis, which is not the case. Likewise, position uncertainties are not included since additional information is necessary for evaluating these measures, and their contribution is negligible for the calculation \cite{8}.

% Then, the simplified equation is:
% \begin{equation}
% S_a= S_h+S_r+S_s
% \label{eq2}
% \end{equation}

Distances can be written in terms of speed and time as:
\begin{equation}
S_a= v_h\left(T_r+T_s\right)+v_r T_r+\frac{v_r T_s}{2}
\label{eq3}
\end{equation}

$v_h$ is the speed at which a human walks towards the robot, defined as $1600$ mm/s in the ISO 13855 standard. $v_r$ is the robot's speed, taken as $500$ mm/s from the UR3 technical specification sheet \cite{11} as 50$\%$ of its nominal speed (i.e., the speed of the robot before it stops). $T_r$ is the robot's reaction time when in the presence of an operator. The latter was determined experimentally using the vision system integrated with the robot's control system, obtaining an average of $28.3$ ms. Finally, $T_s$ is the stop time of the robot, taken as $400$ ms from the UR3 user manual \cite{12}. Therefore, replacing the values above in Eq.~(\ref{eq3}), the minimum separation distance is $S_a~\approx~800$~mm.
%a value for $S_a$ is obtained.

% \begin{equation*}
% \begin{aligned}
% S_a= 1600mm/s\left(0.0283s + 0.400s\right) + \\ 500mm/s\left(0.0283s\right)+\frac{500mm/s\left(0.400s\right)}{2}
% \end{aligned}
% \end{equation*}

% \begin{equation}
% S_a = 799.5 mm \cong 800 mm
% \label{eq4}
% \end{equation}

The result means that if the distance $S$ between the base of the robot and an operator is less than or equal to the MSD,
%($S$ $\leq$ $S_a$) 
the robot must stop. The result is consistent, considering that the maximum reach of the robot is $500$ mm. Although Eq.\ref{eq1} does not consider the robot using an end effector, the maximum reach, including the Robotiq 2F-85, is $662.8$ mm. This length is still within the protection zone, with a margin $\Delta_s = 137.2$ mm between the end effector and the MSD.

%Although Eq.\ref{eq1} does not consider the end effector of the robot, the length of the Robotiq 2F-85 \cite{13} has a maximum reach of 162.8 mm when it is closed. This means that the maximum reach of the robot including the end effector is 662.8 mm, which is also covered within the protection zone with a margin $\Delta_s$ of 137.2 mm between the end effector of the robot and the minimum separation distance 

%(Fig. \ref{ur3_reach}).

% \begin{figure}[htbp]
% \centerline{\includegraphics[scale=.4]{ur3_reach.png}}
% \caption{Maximum reach of the UR3 with the Robotiq 2F-85}
% \label{ur3_reach}
% \end{figure}

Assuming the worst-case scenario, where the robot is at its maximum reach (including its end effector) and an operator is crossing the line of the MSD in a straight line toward the end effector; the estimated collision time between the operator and the end effector is obtained using Eq.~\ref{eq5}.

\begin{equation}
t_{collision}= \frac{\Delta_s}{v_h} = \frac{\SI{137.2}{\mm}}{\SI{1600}{mm/s}} = \SI{0.0858}{s} = \SI{85.8}{ms}
\label{eq5}
\end{equation}

As can be seen, $t_{collision} > T_r$. It indicates that the robot would react to the detection of an operator even in the worst-case scenario. However, during the routine that the robot performs, it never extends to its maximum reach. %It performs a routine within a defined radius of $411.3$ mm, including its end effector.}

\begin{table}[htbp]
\centering
\caption{Boundaries of the safety Zones}
\label{zones_table}
\small
\begin{tabular}{ccc}
\hline
\textbf{Zone} & \textbf{Color} &\textbf{Boundaries} \\ 
\hline
\hline
\textbf{High-risk zone} & \cellcolor{red} & \small\begin{tabular}[c]{@{}c@{}}$S$ $\leq$ $S_a$\\ $S$ $\leq$ 800 mm\end{tabular} \\ \hline
\textbf{Low-risk zone}  & \cellcolor{yellow} & \small\begin{tabular}[c]{@{}c@{}}$S_a$ $<$ $S$ $\leq$ $S_b$\\ 800 mm $<$ $S$ $\leq$ 1550 mm\end{tabular}  \\ \hline
\textbf{Safe zone}     & \cellcolor{green} & \small\begin{tabular}[c]{@{}c@{}}$S$ $>$ $S_b$\\ $S$ $>$ 1550 mm\end{tabular}  \\ \hline
\end{tabular}
\end{table}

The boundaries of the safety zones are summarized in Table \ref{zones_table}. The high-risk zone includes the area from the robot's base to $S_a$. The low-risk zone was defined using the technical specification ISO / TS 15066. It indicates that there must be a clearance of at least $500$ mm of distance. For better visualization of the traffic within that area, a $750$ mm distance was assumed. Therefore, the outer limit of the low-risk zone is defined as $S_b= S_a + 750$ mm $= 1550$ mm. The remaining area is the safe zone, where no safety actions are taken as there is not possibility of collision. This zone is described as any distance S greater than $S_b$.

%The boundaries of the safety zones are summarized in Table \ref{zones_table}. The first zone from the base of the robot to $S_a$ is defined as the high-risk zone (red). The second zone is the low-risk zone (yellow), which is defined using the technical specification ISO / TS 15066. This specification indicates that there must be a clearance of at least $500$ mm of distance. For a better visualization of the traffic within that area, a $750$ mm distance was assumed. Therefore, the outer limit of the low-risk zone is defined as $S_b= S_a + 750$ mm $= 1550$ mm. 

%\ref{eq6}.

% \begin{equation}
% $S_b= S_a + 750 mm = 1550 mm$
% \label{eq6}
% \end{equation}

\begin{figure}[htbp]
\centerline{\includegraphics[scale=.2]{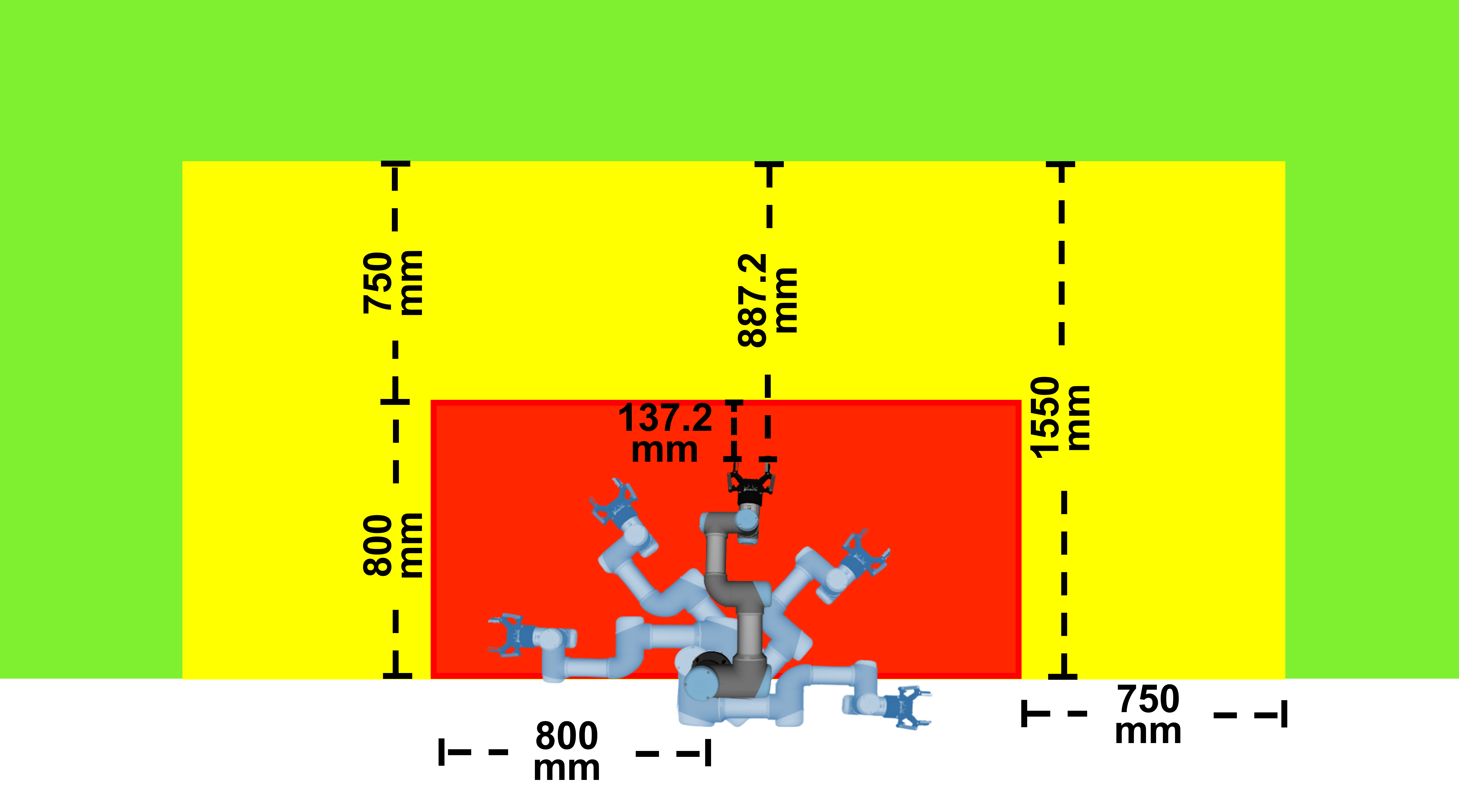}}
\caption{Dimensions of safety zones: high-risk zone (red), low-risk zone (yellow), safe zone (green)}
\label{zones}
\end{figure}

Fig. \ref{zones} shows the dimensions of the zones defined in Table \ref{zones_table} with the robot in the scene. 

\subsection{Vision System}\label{vision_sys}

The vision system handles three processes: operator detection, tracking, and 3D position estimation. The Tensorbox object detection framework \cite{15} was used to detect people from above. This framework was further trained using 21,222 images with similar characteristics. The data was split 50\% for training, 30\% for validation, and 20\% for testing. The algorithm showed an accuracy of 92$\%$ with the test dataset \cite{14}.
The CNN-based detector works in parallel with a tracking algorithm. Therefore, for every frame, the coordinates of the bounding box of the detected heads are sent to the tracker. The tracker uses the received data to predict the operator's position when the detector fails. The Simple Online and Real-time Tracking algorithm (SORT) \cite{16} was used. When tested, it showed a higher accuracy and a higher and steadier tracking speed.   

The safety system requires the operator's position relative to the robot's base and his/her height. This data was estimated by using the intrinsic camera parameters (Kinect \cite{18}), the camera extrinsic parameters (found by calibration), and the camera's depth sensor. The details of the vision system are presented in \cite{14}.

\subsection{Operation Modes}\label{operation_modes}

% After calculating the 3D position of the operator in the scene, and by following the boundaries defined on Table \ref{zones_table}), it is possible to know the safety zone where the operator is located. Fig. \ref{sec_zones} shows the operator in different zones. Being aware of the operator's location will result useful afterwards for the control system, which will change the behavior of the robot, depending on the zone of the closest operator.

% \begin{figure}[h]
% \centerline{\includegraphics[scale=.57]{sec_zones_vision.png}}
% \setlength{\abovecaptionskip}{3pt}
% \setlength{\belowcaptionskip}{-10pt}
% \caption{Detection of the person in different safety zones. The color of the bounding box changes according to the zone color.}
% \label{sec_zones}
% \end{figure}

Knowing which zone the operator is in allows the control system to modify the robot's behavior. Three operation modes for the safety system are proposed, which can be used depending on the needs. For all modes, if there is more than one operator on the scene, the speed will change according to the safety zone of the closest operator. 

\paragraph{SSM with static zones}

%In this mode, the speed of the robot changes according to the zone the person is in, as follows: High-risk zone  0 mm/s,  Low-risk zone  500 mm/s, and Safe zone  1000 mm/s
%Speed changing with static safety zone
%The zones implemented are the ones defined in subsection \ref{safety_zones}, taking the base of the robot as a reference point. The robot's speed will change as indicated in Table \ref{static}.

% \begin{table}[!h]
% \centering
% \caption{Characteristics of the static safety zones}
% \label{static}
% \begin{tabular}{cccc}
% \hline
% \textbf{Zone}           & \textbf{Color} & \textbf{Boundaries} & \textbf{Robot's speed} \\ 
% \hline
% \hline

% \textbf{High-risk zone} & \cellcolor{red}            & $S$ $\leq$ 800 mm   & 0 mm/s          \\ \hline
% \textbf{Low-risk zone}  & \cellcolor{yellow}         & 800 mm $<$ $S$ $\leq$ 1550 mm    & 500 mm/s           \\ \hline
% \textbf{Safe zone}      & \cellcolor{green}          & $S$ $>$ 1550 mm    & 1000 mm/s            \\ \hline
% \end{tabular}
% \end{table}

This mode aims to change the robot's speed according to the operator's relative distance. If operators are detected in the safe zone, the robot moves at 100 $\%$ of its nominal speed, at 50 $\%$ speed if operators are detected in the low-risk zone, and it stops its movement if operators are detected in the high-risk zone. The robot performs a category two protection stop as indicated in the ISO 10218-1:2011 standard. It is a controlled stop in which the robot continues to be powered \cite{12} so it can resume its routine when the operator has left this area.

%This mode aims to change the speed of the robot in accordance to the operator's distance relative to the robot. When an operator is detected in the safe zone, the robot will move at 100 $\%$ of its nominal speed ($1000$ mm/s), if they move to the low-risk zone, the robot will decrease its speed by 50 $\%$, and if they enter the high-risk zone, the robot will stop its movement. The opposite occurs if the operator moves from the high-risk zone to the safe zone. The behavior of the robot when an operator is detected in the high-risk zone, is to perform a category 2 protection stop as indicated in the ISO 10218-1:2011 standard. This is a controlled stop in which the robot continues powered \cite{12}, so that it can resume its routine when the operator has left this area.

\paragraph{SSM dynamic zone}

\begin{figure}[h]
\centerline{\includegraphics[scale=.21]{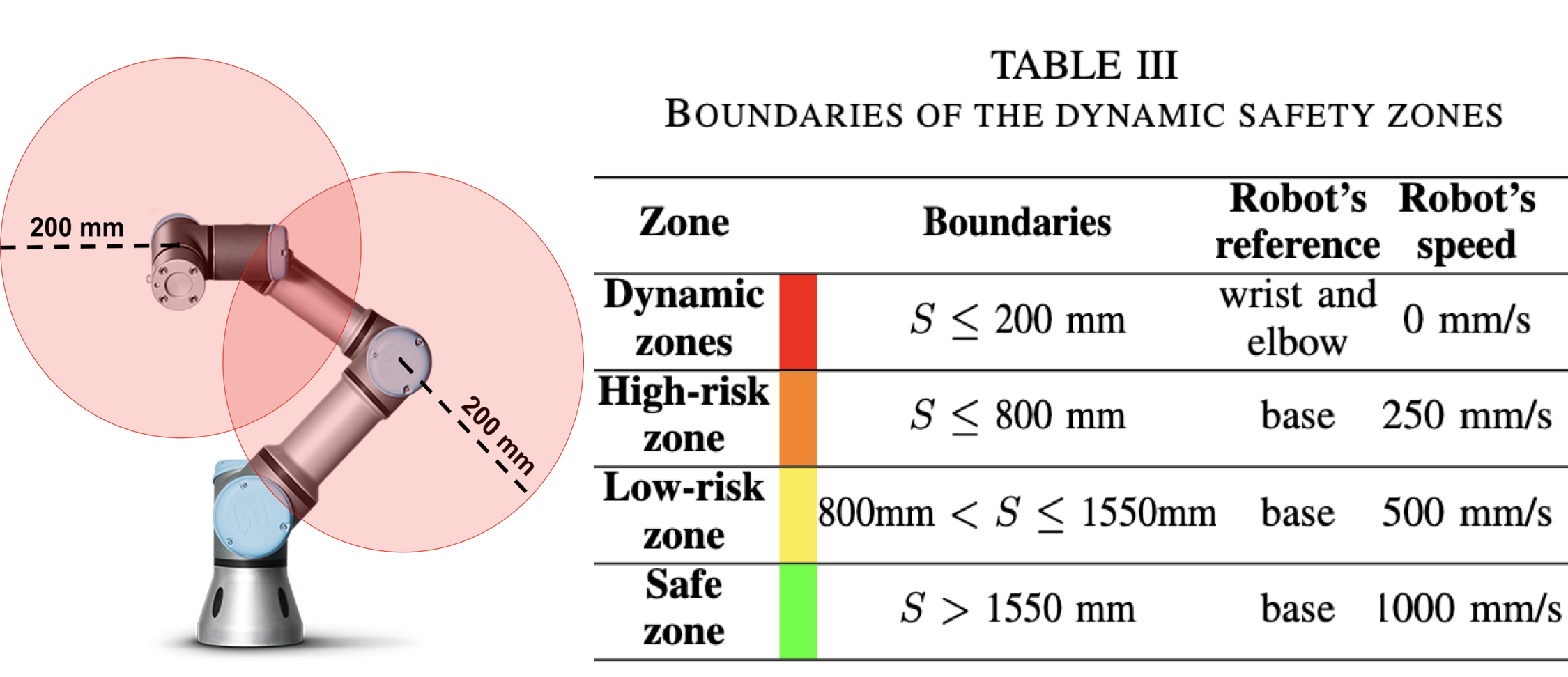}}
\caption{Safety zones around wrist and elbow joints. By controlling these zones, the robot will stop only when required, ensuring safety and efficiency.}
\label{dynamic_zones}
\end{figure}

Compared to the previous mode, the robot does not stop its movement if the operator enters the high-risk zone. Instead, an additional analysis is performed to stop the robot only when it is strictly necessary to ensure the operator's safety and greater efficiency in the robot's task. To achieve this, dynamic safety zones are created around those critical joints of the robot that can cause a collision with the operator. Fig. \ref{dynamic_zones} shows the new safety zones created around the wrist and elbow joints. Those joints were chosen since they are the ones that move along the Cartesian plane.

By inspecting the surroundings of these joints as the robot moves, it is possible to have a closer human-robot interaction and to stop the robot only if the operator is close enough to collide with the robot. The MSD between the operator and any of the two joints was established equal to $200$ mm as shown in Fig. \ref{dynamic_zones}, considering the collision time calculated in Eq. \ref{eq5} and covering all moving parts of the robot.  

Since the protection zones change their position as the robot moves, this operation mode offers a freer human-robot interaction, maintaining a minimum safety distance between them. Table III presents the boundaries of the dynamic safety zones. The robot only stops if the horizontal distance between the detected operator and the robot's wrist or elbow is less than or equal to $200$ mm. Like in the previous mode, a category 2 protection stop is performed, so that the robot can resume its routine when the operator walks away from it.

\paragraph{Obstacle avoidance}
This operation mode seeks to replace the robot's protection stop with an obstacle avoidance algorithm, such that the robot does not stop its routine at all, and the operator's safety is still ensured. In this mode, %the robot does not stop if an operator enters the high-risk zone. Instead, 
the speeds for the safe and low-risk safety zones work the same as in the previous mode. However, if an operator enters the high-risk zone, the robot will move at 10$\%$ of its nominal speed. The latter allows the vision system to detect the operator in real-time, i.e., an obstacle in the robot's workspace, and the robot to change its trajectory to avoid the obstacle. The latter is achieved with the \textit{allow$\_$looking} and  \textit{allow$\_$replanning} functions from MoveIt \cite{19} (that were inactive for previous modes). They allow the robot to search for alternate paths to reach its destination point as it moves.

\begin{figure}[h]
\centerline{\includegraphics[scale=.23]{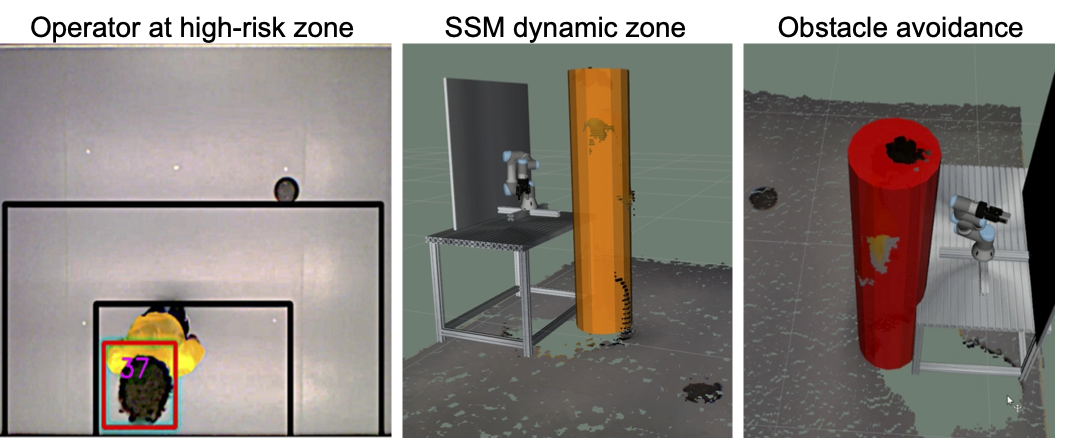}}
\caption{Speed and Separation Monitoring (SSM) results when an operator enters the high-risk safety zone (left image).}
\label{simulation2}
\end{figure}

Fig. \ref{simulation2} shows two operation modes in action. When the operator enters the high-risk zone (left image), if the SSM dynamic mode is active, an orange cylinder is drawn in the simulated workspace for the robot to be aware of the operator's location (middle image). At this point, the robot's speed is reduced to 25$\%$ of its nominal speed. On the other hand, the figure on the right shows the obstacle avoidance operation mode. The red cylinder is drawn in the simulated workspace, and the robot's speed is reduced to 10$\%$ of its nominal speed. 

%In parallel, the allow$\_$looking and allow$\_$replanning functions are being executed as seen in the upper left terminal. On the right side, it is observed that the operator interferes with the robot's path, therefore, the robot changes its trajectory to avoid the operator, who is seen as an obstacle.

\section{Experiments and Results}
Three experiments were carried out (all in the SSM operation mode with static zones) to evaluate the performance of the safety system and how it is affected by the number of people on the scene. The first experiment evaluates the system's ability to classify the detected people on the scene correctly. The second and third experiments evaluate the robot's reaction and stop time when the operator moves from one safety zone to another and when he/she enters the high-risk zone. Videos are available in \footnote{Videos of the safety system \url{https://youtube.com/playlist?list=PLgYA51rB9xXw1RwwL0Udv1Ikwel68HY3y}}

\subsection{Risk Zone Identification}
This experiment determines how accurate the vision system classifies the zone where the operator is detected compared with the actual zone. A video (900 frames) where operators walked around the scene was recorded, and the operators' heads were manually labeled to obtain ground truth data. Table~\ref{zone_class} shows the different performance measurements calculated per  zone.

%A confusion matrix was made for each zone and with this information,Different performance measurements were calculated and summarized in Table~\ref{zone_class}.

%With a correct classification of the occupied areas, the correct speed change of the robot is obtained, ensuring the operator's safety

\begin{table}[htbp]
\setcounter{table}{3}
\centering
\caption{Performance of the system's zone classifier}
\label{zone_class}
\small
\begin{tabular}{cccc}
\hline
\textbf{Measure}   & \textbf{Green zone} & \textbf{Yellow zone} & \textbf{Red zone} \\ 
\hline
\hline
\textbf{Accuracy}  & 85.6 $\%$           & 91.9 $\%$            & 97.9 $\%$         \\ \hline
\textbf{Precision} & 97.7 $\%$           & 99.4 $\%$            & 99.5 $\%$         \\ \hline
\textbf{Recall}    & 57.4 $\%$           & 81.9 $\%$            & 91.5 $\%$         \\ \hline
\textbf{F-score}   & 72.3 $\%$           & 89.8 $\%$            & 95.3 $\%$         \\ \hline
\end{tabular}
\end{table}

%Table~\ref{zone_class} shows that 
The accuracy and precision of the zone classifier were greater than 85 $\%$ and 97 $\%$ respectively for each of the three zones. Overall, the system correctly classifies the safety zone where the operator was located. A low recall percentage was obtained for the green zone, indicating that only 57.4 $\%$ of the cases were detected correctly. Most false negatives are caused by the absence of detections and not by a wrong classification. The latter can be attributed to the Kinect’s disparity effect, which reduces the usable field of view (no depth data on the borders of the image). Therefore, the number of correct detections in the green zone is reduced, and the recall score decreases.

\subsection{Reaction time}
This experiment calculates the time it takes for the robot to change its speed from when the system sends the information that an operator has moved from the safe zone to the low-risk zone or vice versa. The average reaction time is calculated after running the system for 1 minute in real-time while operators were walking between both zones. The results are compared with the reference reaction time ($T_r$) calculated in Section~\ref{def_zones}.

\begin{figure}[h]
\centerline{\includegraphics[scale=.14]{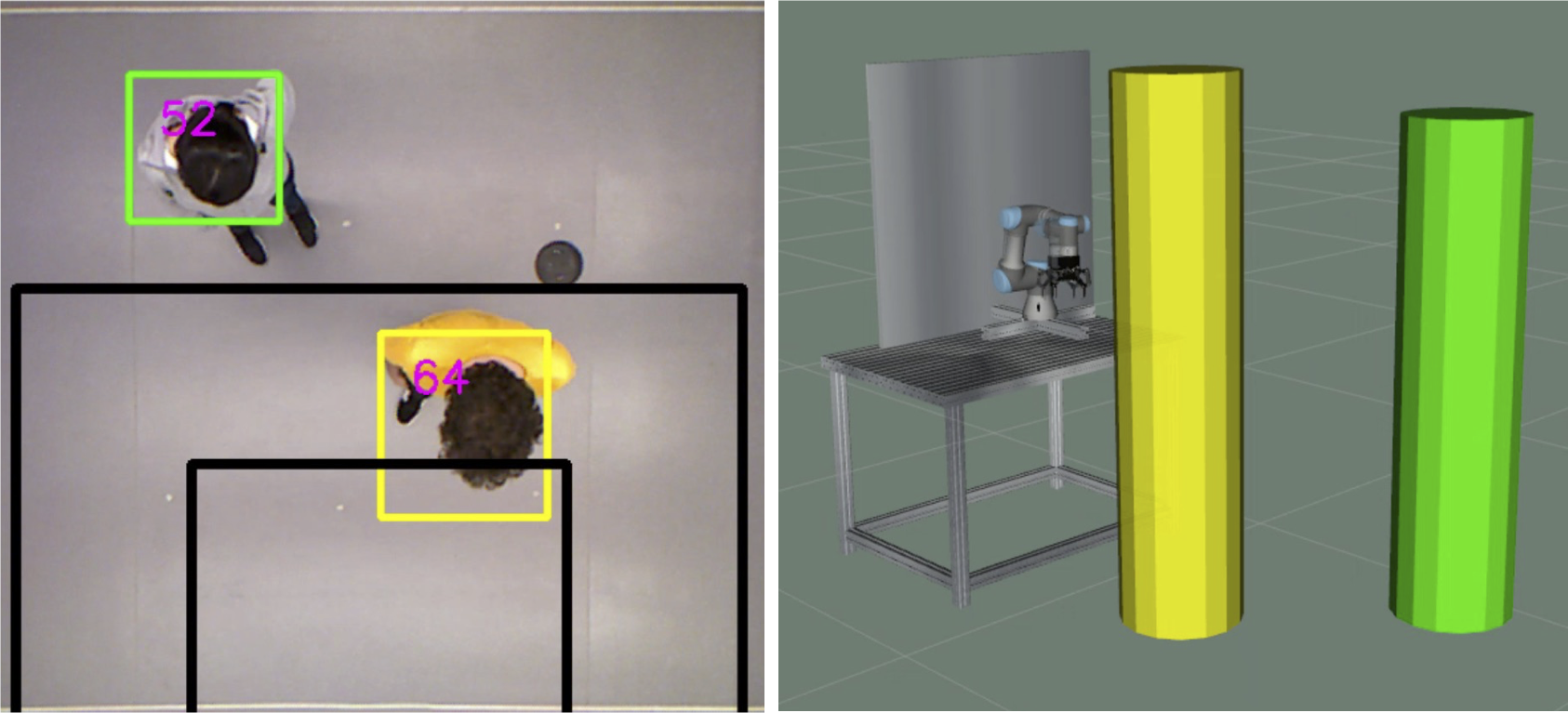}}
\caption{Reaction time tests. Reaction time tests. Results of the vision system (left) and the way operators are perceived as obstacles by the robot (right).}
\label{multiple}
\end{figure}

Table~\ref{reaction_table} shows the average reaction time for two tests carried out with one and two people in the scene. For each case, 20 zone changes were made, in which the operators switched from the safe zone to the low-risk zone or vice versa. 
Fig. \ref{multiple} shows an example of the experiment. The image on the left shows the results of the vision system. The image on the right shows how the operators are perceived as obstacles by the robot when they are represented as 3D figures.

\begin{table}[htbp]
\centering
\caption{Average reaction time between safe zone and low-risk zone}
\label{reaction_table}
\small
\begin{tabular}{cccc}
\hline
\multirow{2}{*}{\textbf{\small\begin{tabular}[c]{@{}c@{}}\# People\\ in the scene\end{tabular}}} & \multicolumn{3}{c}{\textbf{Average $t_r$ (ms)}} \\ \cline{2-4} 
           & \multicolumn{1}{c}{\textbf{Test 1}} & \multicolumn{1}{c}{\textbf{Test 2}} & \textbf{Average} \\ \hline
           \hline
\textbf{1} & \multicolumn{1}{c}{38.3}            & \multicolumn{1}{c}{32.8}            & 35.6             \\ \hline
\textbf{2} & \multicolumn{1}{c}{54.5}            & \multicolumn{1}{c}{52.5}            & 53.5             \\ \hline
\end{tabular}
\end{table}

Results registered in Table~\ref{reaction_table} show that by increasing the number of people in the scene from one to two, $t_r$ %the reaction time 
increased by an average of 17.9 ms. This increment is coherent because more 3D figures must be created in the simulation environment, and more calculations are required to find the closest operator to the robot. Nevertheless, the found values do not represent a risk to people passing through the scene. If we consider that the time it takes an operator to cross the entire low-risk zone in a straight line ($750$ mm) at an average speed of $1600$ mm/s (Section~\ref{def_zones}) is about 469 ms, the different $t_r$ %reaction times 
found in Table~\ref{reaction_table} are sufficient for the robot to change its speed, even when multiple operators are on the scene.

On the other hand, the values in Table~\ref{reaction_table} are greater than the reference $T_r$ %reaction time 
calculated in Section~\ref{def_zones} ($T_r$ = 28.3 ms). This increment occurs because $T_r$ was estimated without considering the system's underlying calculations, such as zone classification and generation of 3D collision objects in the simulated environment.

\subsection{Stop time}
This experiment calculates the time it takes for the robot to stop its movement from the moment the system sends the information of the operator detected in the high-risk zone. The average $t_s$ %stop time 
is calculated after running the system for 1 minute in real-time while operators were entering the high-risk zone. The results are compared with the reference $T_s$ %stop time 
($T_s$ = 400 ms) discussed in Section~\ref{def_zones}, and the estimated $t_{collision}$ %collision time
calculated in Eq.\ref{eq5}.

Table \ref{stop_table} shows the averages of $t_s$ %stop times 
for two tests carried out with one and two people in the scene. For each case, 20 zone changes were made, in which the operators walked from the low-risk zone to the high-risk zone.

\begin{table}[htbp]
\centering
\caption{Average stop time when entering the high-risk zone}
\label{stop_table}
\small
\begin{tabular}{cccc}
\hline
\multirow{2}{*}{\textbf{\small\begin{tabular}[c]{@{}c@{}}\# People\\ in the scene\end{tabular}}} & \multicolumn{3}{c}{\textbf{Average $t_s$ (ms)}} \\ \cline{2-4} 
           & \multicolumn{1}{c}{\textbf{Test 1}} & \multicolumn{1}{c}{\textbf{Test 2}} & \textbf{Average} \\ \hline
           \hline
\textbf{1} & \multicolumn{1}{c}{67.1}            & \multicolumn{1}{c}{60.2}            & 63.7             \\ \hline
\textbf{2} & \multicolumn{1}{c}{76.9}            & \multicolumn{1}{c}{75.5}            & 76.2             \\ \hline
\end{tabular}
\end{table}

Similar to the previous test, when the number of people in the scene increases, $t_s$ %the stop time 
increases too, due to the extra computational cost of having two operators in the scene.

The times obtained in these tests are smaller than the reference $T_s$ of the UR3. This difference is because $T_s$ was estimated considering that the robot is stopped through the immediate disconnection of its power. While for our safety system, a control loop is started to maintain the robot in the last position \cite{12}. This action requires less time to stop the robot than waiting for it to be entirely powered off.

It is also observed that the different $t_s$ found
%stop times 
are lower than the calculated $t_{collision}$ %collision time 
($t_{collision}$ = 85.8 ms) in the worst-case scenario presented in Section~\ref{def_zones} %that is when the robot is fully extended, and a person is heading towards it in a straight line. 
Therefore, it can be guaranteed that the robot will stop in time, and there will be no collision between the operator and the robot.

\section{Conclusions and Future Work}
This paper presented a safety system for an industrial manipulator, combining Speed and Separation Monitoring (SSM) operation with computer vision and deep learning techniques. The results from the experiments demonstrated that the system's performance ensures the operator's safety within the robot's workspace. People were classified in the high-risk zone with an accuracy of 97.9$\%$ and a precision of 99.5$\%$. In the second and third experiments, the average $t_r$ and $t_s$ were 53.5 ms and 76.2 ms for two operators in the scene. Both values, when compared to the ones found for the definition of the safety zones, validated that the robot reacts and stops within the time limits when the operator enters the high-risk zone. The three operation modes proposed in the paper introduced distinct types of human-robot interaction. The results confirmed that it is feasible to eliminate physical barriers around an industrial robot while keeping the operator's safety. Future improvements are guided to include kinematic and dynamic models of the robot and its operator in the safety system to anticipate their behavior and execute more complex actions.

\vspace{12pt}

\begin{thebibliography}{00}

\bibitem{I1} Universal Robots. (n.d.). Historia de los cobots. Retrieved June 1, 2020, from https://www.universal-robots.com/es/acerca-de-universal-robots/noticias/historia-de-los-cobots/
\bibitem{I2} Beaupre, M. (2014). Collaborative Robot Technology and Applications [Slides]. Retrieved from https://www.robotics.org/userAssets/riaUploads/file/4-KUKA\_Beaupre.pdf
\bibitem{I3} ISO. (2016). ISO/TS 15066:2016(en) Robots and robotic devices — Collaborative robots. Retrieved May 20, 2020, from https://www.iso.org/obp/ui/\#iso:std:iso:ts:15066:ed-1:v1:en 

\bibitem{1} Malm, T., Salmi, T., Marstio, I., \& Montonen, J. (2019). Dynamic safety system for collaboration of operators and industrial robots. Open Engineering, 9(1), 61–71. https://doi.org/10.1515/eng-2019-0011
\bibitem{2} Mohammed, A., Schmidt, B., \& Wang, L. (2016). Active collision avoidance for human–robot collaboration driven by vision sensors. International Journal of Computer Integrated Manufacturing, 30(9), 970–980. https://doi.org/10.1080/0951192x.2016.1268269
\bibitem{3} Robla-Gomez, S., Becerra, V. M., Llata, J. R., Gonzalez-Sarabia, E., Torre-Ferrero, C., \& Perez-Oria, J. (2017). Working Together: A Review on Safe Human-Robot Collaboration in Industrial Environments. IEEE Access, 5, 26754–26773. https://doi.org/10.1109/access.2017.2773127
\bibitem{4} ISO. (2011). ISO 10218-2:2011(en) Robots and robotic devices — Safety requirements for industrial robots — Part 2: Robot systems and integration. Retrieved May 20, 2020, from https://www.iso.org/obp/ui/\#iso:std:iso:10218:-2:ed-1:v1:en

\bibitem{6} ABB. (2020). ABB Líder en tecnologías digitales para la industria. Retrieved June 8, 2020, from https://new.abb.com/south-america
\bibitem{6.1} B. Matthias, S. Kock, H. Jerregard, M. Kallman, I. Lundberg and R. Mellander, "Safety of collaborative industrial robots: Certification possibilities for a collaborative assembly robot concept," 2011 IEEE International Symposium on Assembly and Manufacturing (ISAM), Tampere, 2011, pp. 1-6, doi: 10.1109/ISAM.2011.5942307.
\bibitem{7} Medina, A., Mora, J., Martinez, C., Barrero, N., \& Hernandez, W. (2019). Safety Protocol for Collaborative Human-Robot Recycling Tasks. IFAC-PapersOnLine, 2008–2013. https://doi.org/10.1016/j.ifacol.2019.11.498 
\bibitem{8} Byner, C., Matthias, B., \& Ding, H. (2019). Dynamic speed and separation monitoring for collaborative robot applications – Concepts and performance. Robotics and Computer-Integrated Manufacturing, 239–252. https://doi.org/10.1016/j.rcim.2018.11.002
\bibitem{9} Marvel, J., \& Norcross, R. (2017). Implementing speed and separation monitoring in collaborative robot workcells. Robotics and Computer-Integrated Manufacturing, 144–155. https://doi.org/10.1016/j.rcim.2016.08.001
%\bibitem{10} Belingardi, G., Heydaryan, S., \& Chiabert, P. (2017). Application of speed and separation monitoring method in human-robot collaboration: industrial case study. 
\bibitem{11} Universal Robots. (2016). UR3 Especificaciones técnicas. Retrieved June 3, 2020, from https://www.universal-robots.com/media/240781/ur3\_sp.pdf
\bibitem{12} Universal Robots. (2018b). UR3 user manual. Retrieved June 2, 2020, from https://s3-eu-west-1.amazonaws.com/ur-support-site/32341/UR3\_User\_Manua=l\_en\_E67ON\_Global-3.5.5.pdf
%\bibitem{13} Robotiq. (2018, November 7). Robotiq 2F-85 \& 2F-140 for e-Series Universal Robots. Retrieved May 29, 2020, from https://assets.robotiq.com/website-assets/support\_documents/document/2F-85\_2F-140\_Instruction\_Manual\_e-Series\_PDF\_20190206.pdf 
\bibitem{14} Duque-Suárez, N., Amaya-Mejía, L. M., Martinez, C., \& Jaramillo-Ramirez, D. (2021). Deep Learning for Safe Human-Robot Collaboration. Advances in Automation and Robotics Research, 239–251. https://doi.org/10.1007/978-3-030-90033-5\_26
\bibitem{15} Stewart, R., Andriluka, M., Ng, A. (2016). End-to-End People Detection in Crowded
Scenes. 2016 IEEE Conference on Computer Vision and Pattern Recognition (CVPR).
https://doi.org/10.1109/cvpr.2016.255
\bibitem{16} Bewley, A., Ge, Z., Ott, L., Ramos, F., Upcroft, B. (2016). Simple online and re-
altime tracking. 2016 IEEE International Conference on Image Processing (ICIP).
https://doi.org/10.1109/icip.2016.7533003

\bibitem{17} Mallick, S. (2017, February 13). Object Tracking using OpenCV (C++/Python). Retrieved May
23, 2020, from https://www.learnopencv.com/object-tracking-using-opencv-cpp-python/

\bibitem{18} Al-Naji, AA, Gibson, K, Lee, S-H, Chahl, J. (2017). Real Time Apnoea Monitoring of Children
Using the Microsoft Kinect Sensor: A Pilot Study.Sensors.17.286.10.3390/s17020286.


\bibitem{19} ROS. (n.d.). MoveIt Tutorials. Retrieved May 21, 2020, from https://ros-planning.github.io/moveit\_tutorials/
\bibitem{20} Kavraki Lab, Department of Computer Science, Rice University. (2020). The Open Motion Planning Library. Kavrakilab. http://ompl.kavrakilab.org/

%\bibitem{21} Videos.  \url{https://youtube.com/playlist?list=PLgYA51rB9xXw1RwwL0Udv1Ikwel68HY3y}

\end{thebibliography}
\end{document}